\documentclass[letterpaper, 10 pt, journal, twoside]{IEEEtran}
\usepackage{amsmath,amsfonts}
\usepackage{algorithmic}
\usepackage{algorithm}
\usepackage{array}
\usepackage[caption=false,font=footnotesize]{subfig}
\usepackage[font=footnotesize]{caption}
\usepackage{textcomp}
\usepackage{stfloats}
\usepackage{verbatim}
\usepackage{graphicx}
\usepackage[sort&compress,numbers,sectionbib]{natbib}
\usepackage{booktabs}

\usepackage[hyphens]{url}

\usepackage{hyperref}
\hypersetup{bookmarksopen,bookmarksnumbered,
pdfpagemode=UseOutlines,
colorlinks=true,
linkcolor=teal,
anchorcolor=teal,
citecolor=teal,
filecolor=teal,
menucolor=teal,
urlcolor=teal,
breaklinks=true
}
\usepackage{color,soul}
\usepackage{svg}

\begin{document}

\title{Characterizing the Complexity of Social Robot Navigation Scenarios}


\author{Andrew Stratton$^{1}$, Kris Hauser$^{2}$, and Christoforos Mavrogiannis$^{1}$%
\thanks{A. Stratton was partially funded by NSF Grant \# NRI-2025782.} 
\thanks{$^{1}$Andrew Stratton and Christoforos Mavrogiannis are with the Department of Robotics, University of Michigan, Ann Arbor, USA.
        {\tt\footnotesize \{arstr, cmavro\}@umich.edu.}}%
\thanks{$^{2} $Kris Hauser is with the Department of Computer Science, University of Illinois at Urbana-Champaign.
        {\tt\footnotesize kkhauser@illinois.edu.}}%
}


\markboth{IEEE Robotics and Automation Letters. Preprint Version. Accepted November, 2024}
{Stratton \MakeLowercase{\textit{et al.}}: Characterizing Complexity} 


\maketitle

\begin{abstract}
Social robot navigation algorithms are often demonstrated in overly simplified scenarios, prohibiting the extraction of practical insights about their relevance to real-world domains. Our key insight is that an understanding of the inherent complexity of a social robot navigation scenario could help characterize the limitations of existing navigation algorithms and provide actionable directions for improvement. Through an exploration of recent literature, we identify a series of factors contributing to the complexity of a scenario, disambiguating between contextual and robot-related ones. We then conduct a simulation study investigating how manipulations of contextual factors impact the performance of a variety of navigation algorithms. We find that dense and narrow environments correlate most strongly with performance drops, while the heterogeneity of agent policies and directionality of interactions have a less pronounced effect. Our findings motivate a shift towards developing and testing algorithms under higher-complexity settings.
\end{abstract}

\begin{IEEEkeywords}
Human-Aware Motion Planning, Human-Centered Robotics, Autonomous Vehicle Navigation.
\end{IEEEkeywords}

\section{Introduction}
\IEEEPARstart{R}{ecent} surveys on social robot navigation~\citep{Mavrogiannis2021CoreCO, Francis2023PrinciplesAG, singamaneni2023survey} have highlighted that strong assumptions in typical evaluation practices prohibit transfer to real-world domains. Much of the literature considers sparse, slowly navigating human crowds, giving rise to non-interactive human-robot encounters that are easily handled by classical navigation algorithms. Hence, the benefits of many modern approaches are not transparent. This underscores the need for standardized benchmarks; however, defining benchmarks for this domain requires community consensus over appropriate experimental design. This would need to capture several aspects, such as the spatial arrangement and roles of agents, the briefing that human users receive before the experiment, robot design specifications, metrics evaluating robot performance and human impressions, etc.

\begin{figure}[t!]
  \includegraphics[width=\linewidth]{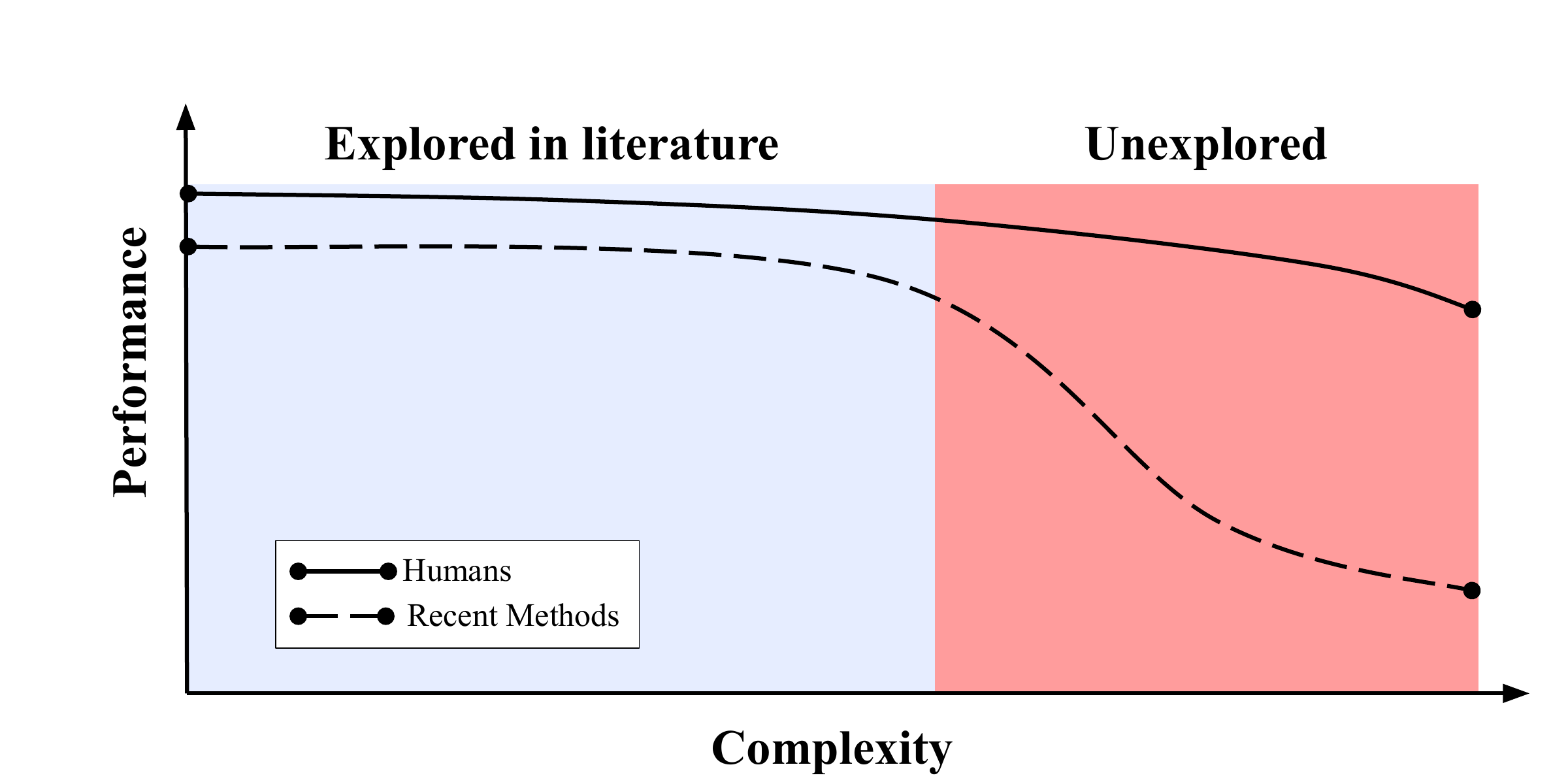}
  \caption{Humans can seamlessly handle a wide range of crowd navigation scenarios~\citep{dendorfer2019cvpr19trackingdetectionchallenge}. In contrast, social robot navigation algorithms struggle to handle scenarios with realistic levels of Complexity \citep{trautmanijrr}. In this paper, we show that high-Complexity scenarios are underexplored in the literature. We demonstrate principal factors contributing to poor navigation performance, and argue for the definition of benchmarks that account for their impact.}
  \vspace{-5mm}
  \label{fig:summary}
\end{figure}

In this paper we argue that to establish effective benchmarks for social robot navigation, the research community should better understand and control the \emph{dimensions of problem Complexity}. Clear definitions of Complexity in fields like theoretical computer science~\citep{arora2006computational}, learning theory~\citep{vapnik1998statistical}, and motion planning~\citep{farber2001topological} enable a characterization of limitations in existing algorithms and motivate future research directions. However, robotics is a synthetic science~\citep{koditschek2021robotics}, combining many elements from other fields, which makes it harder to formulate similarly succinct and relevant definitions for Complexity. Social robot navigation~\citep{Mavrogiannis2021CoreCO} is no exception to this complication, as it lies at the intersection of multiple research areas, including motion planning, machine learning, control, and human modeling, among others. In this work, we approach Complexity through an empirical lens: a dimension of {\em Complexity} is a factor that causes a drop in {\em Performance} of an automated method (Fig.~\ref{fig:summary}). Here, Performance is measured with respect to relevant criteria accepted by the community~\citep{Mavrogiannis2021CoreCO}.

To instantiate our framework, we review the evaluation practices of recent literature and extract a set of factors contributing to the Complexity of a social robot navigation scenario, distinguishing between contextual and robot-related factors. We then conduct an extensive simulated study to understand how robot-independent, contextual factors impact algorithmic performance. We find that Density and Environment Geometry have the strongest correlation with performance decline across a wide range of algorithms. We also find that the scenarios most frequently explored in prior work tend to be of lower Complexity, and allow simple, reactive methods to perform effectively. Our findings underscore the need for algorithms that explicitly account for diverse, high-Complexity scenarios, and benchmarks that explore performance in scenarios that have high Complexity across multiple factors.

\section{Related Work}

Recent surveys on social robot navigation~\citep{Mavrogiannis2021CoreCO, Francis2023PrinciplesAG, singamaneni2023survey, gao-eval} emphasize the need for a standardization of benchmarking practices. \citet{Mavrogiannis2021CoreCO} observe the lack of rigorous statistical characterization of robot performance and human impressions, and emphasize the need for better repeatability. \citet{Francis2023PrinciplesAG} collect a list of principles that a social navigation system should take into account, and propose guidelines for evaluating them. \citet{gao-eval} delve into the evaluation methodologies adopted by researchers and practitioners. \citet{singamaneni2023survey} provide a taxonomy of the literature, capturing aspects such as the robot type, the algorithmic approaches used, and the social navigation context. This work complements prior surveys by analyzing the Complexity of the scenarios employed in the evaluation practices of published approaches. By understanding the explored territory of Complexity, we provide directions for integrating more Complex settings in benchmarking practices that more closely resemble what a robot would experience in the real world.

A major challenge in standardizing benchmarking in social navigation lies in the design of experiments, and specifically in balancing repeatability and emergence of natural interactions. \citet{Mavrogiannis2021CoreCO} classify real-world experiments into three classes: demonstrations~\citep{tai2018socially, liu2020decentralized, liu2023intention}, lab studies~\citep{Mavrogiannis18, mavrogiannis2023winding, poddar2023crowd, pacchierotti_eval, Lo19}, and field studies~\citep{trautmanijrr, minerva-robot, rhino-robot}, advocating for the extraction of statistical insights across social navigation experiments. However, the cost of conducting real-world experiments has motivated experimentation within virtual environments~\citep{biswas2021socnavbench}. Borrowing tools from graphics and crowd simulation~\citep{ORCA,Helbing95}, several works have developed photorealistic simulators~\citep{holtz2022socialgym, biswas2021socnavbench, sean}, while others have focused on finding quantitative metrics that capture aspects of performance previously only understood through qualitative metrics~\citep{walker2021corl,social-momentum-thri,singamaneni_bench}. However, as demonstrated by~\citet{fraichard2020}, simulation is often introduced in evaluation practices with shortcomings. Common assumptions in simulated evaluations include the robot being \emph{invisible} to humans, humans being simulated as homogeneous overly submissive agents, the scenarios considered being generated at random, and the scenarios containing no walls or obstacles. Such assumptions result in surprising observations, such as that a ``blind'' robot moving straight can outperform state-of-the-art agents~\citep{Mavrogiannis2021CoreCO}. Motivated by these observations, we contribute an experimental design that gives rise to more realistic crowd conditions by considering the manipulation of contextual variables including the Environment Geometry, the agent Policy Mixture, the crowd Directionality, and Density.

Another challenge in designing benchmarks is the lack of understanding of what makes a social navigation scenario Complex. In prior work, the crowd Density~\cite{trautmanijrr, Mavrogiannis2021CoreCO, sun2021trajectories} has often been used as a proxy for Complexity. However, Density on its own does not capture the Complexity of the motion coupling between closely interacting agents. To capture that, metrics like Path Irregularity~\citep{guzzi_pi} and the Topological Complexity Index~\citep{dynnikov:hal-00001267, social-momentum-thri} quantify aspects of geometric and topological richness of agents' interactions. Our work is relevant to these approaches, however it seeks to extract a more fundamental understanding of how the parameters defining a social navigation scenario impact its Complexity. We are inspired by the Complexity definitions of theoretical fields~\citep{arora2006computational,vapnik1998statistical,farber2001topological}. While our investigation is empirical, we see it as a first step towards defining a formal representation of Complexity.
\section{Factors of Complexity in the Literature} 

We first identified factors of Complexity in the literature by investigating benchmarking practices in papers from the ICRA, IROS, CoRL, and RSS conferences from 2015 to 2024, using the keywords ``social'' and ``crowd''. Additionally, we reviewed work cited in recent surveys~\citep{Mavrogiannis2021CoreCO,Francis2023PrinciplesAG,singamaneni2023survey,gao-eval}. We studied a variety of evaluation methodologies and found a helpful breakdown of the factors characterizing the evaluation of each scenario could be through two categories: contextual (see Fig.~\ref{fig:exp_configurations}) and robot-related factors.  We tabulate the results in Table~\ref{tab:literature}.

\label{sec:factors-literature}

\begin{table*}[ht]
\centering
\caption{Complexity factors in the literature. Directionality: Passing (P), Crossing (C), Circle Crossing (CC), Random (R); Density: Low $[0, 0.1)$, Medium $[0.1, 0.25]$, High $(0.25, \infty)$ $agents/m^2$; Environment: Small room $[0, 25)$, Medium room $[25, 100]$, Large room $(100, \infty)$ $m^2$.\label{tab:complexity}}
\begin{scalebox}{.94}
{
\begin{tabular}{lllllllllll}
  \toprule
Paper & Density &  Directionality & Environment & Robot Size $(mm^3)$ & $v_{max} (m/s)$ & Kinematics & Sensors & Viewpoint \\ 
  \midrule
\citet{angelopoulos_nonverbal} & Low & P & Hallway & 425x480x1210 & 0.85 & DD & N/A & Egocentric \\
\citet{Bennewitz05} & Low & N/A & Office & 520x520x1180 & 0.65 & N/A & Stereo, LIDAR & Egocentric \\
\citet{rhino-robot} & Medium & N/A & Museum & N/A & 0.80 & N/A & LIDAR & Egocentric \\
\citet{ciou_composite} & Low & P & Hallway & 381x455x217 & 1.20 & DD & LIDAR, RGB-D & Egocentric \\
\citet{chen-icra17} & Medium & P, C & Hallway & N/A & N/A & DD & N/A & N/A \\
\citet{katyal_group} & Low & P & Med. Room & 1100x500x610 & 1.60 & Quad & Motion Capture & Overhead \\
\citet{Kim2016} & N/A & P & Med. Room & N/A & N/A & N/A & RGB-D & Egocentric \\
\citet{kirby_thesis} & Low & P & Hallway & 520x520x1180 & 0.70 & N/A & Laser scanner & Egocentric \\
\citet{liu2023intention} & Medium & N/A & Med. Room & 354x354x420 & 0.50 & DD & LIDAR & Egocentric \\
\citet{liu_drl} & Medium & P, C & Outdoor & N/A & N/A & DD & LIDAR & Egocentric \\
\citet{liu2020decentralized} & Low & P, R & Med. Room & 354x354x420 & 0.65 & DD & RGB-D & Egocentric \\
\citet{Lo19} & Low & P & L. Room & N/A & 1.50 & Omni & Motion Capture & Overhead \\
\citet{matsuzaki_dwa} & Medium & P, C & L. Room & N/A & 1.20 & LIDAR & N/A & Egocentric \\
Matsuzaki et al. \cite{matsuzaki_ddrl} & High & P, C & L. Room & N/A & 0.50 & N/A & N/A & N/A \\
\citet{mavrogiannis2023winding} & Medium & P, C & Sm. Room & N/A & 0.75 & 
Omni & Motion Capture & Overhead \\
\citet{social-momentum-thri} & Medium & P, C & Med. Room & 508x660x1580 & 0.83 & Omni & Motion Capture & Overhead \\
\citet{mun2023occlusionaware} & Low & P & L. Room & 354x354x420 & 0.65 & DD & LIDAR & Egocentric \\ 
\citet{mustafa2023probabilistic} & Low & P & Med. Room & 504x430x250 & 2.00 & DD & N/A & N/A \\
\citet{nguyen2023humanlike} & Low & P & Hallway & 810x320x640 & 4.80 & Ackermann & LIDAR & Egocentric \\
\citet{oh_scan} & Low & P, C & L. Room & 504x430x250 & 2.00 & DD & LIDAR, Stereo & Egocentric \\
\citet{pacchierotti_eval} & Low & P & Hallway & N/A & 0.60 & N/A & LIDAR & Egocentric \\
\citet{paezgranados2022pedestrianrobot} & High & P, C & Outdoor & N/A & N/A & N/A & LIDAR, RGB-D & Egocentric \\
\citet{peddi_data} & Medium & P & Sm. Room & 960x793x296 & 0.50 & Omni & Motion Capture & Overhead \\
\citet{poddar2023crowd} & Medium & P, C & Sm. Room & N/A & 1.50 & Omni & Motion Capture & Overhead \\
\citet{qin_dil} & Low & P, C & L. Room & N/A & 0.85 & DD & LIDAR & Egocentric \\ 
\citet{qiu2022learning} & Medium & P, C & Hallway & 354x354x420 & 0.65 & DD & LIDAR, RGB-D & Egocentric \\ 
\citet{sathyamoorthydensecavoid} & High & N/A & Hallway & 354x354x420 & 1.00 & DD & LIDAR, RGB-D & Egocentric \\
\citet{shiomi_towards} & N/A & N/A & L. Room & 600x600x1200 & 0.75 & N/A & LIDAR & Overhead \\
\citet{silva_online} & Low & P & Med. Room & 425x480x1210 & 0.55 & DD & Motion Capture & Overhead \\ 
\citet{singamaneni2022watch} & Low & P & Doorway & 668x668x1640 & 1.00 & N/A & LIDAR & Egocentric \\
\citet{tai2018socially} & Low & P, C & M. Room & 281x306x141 & 0.26 & DD & LIDAR, RGB-D & Egocentric \\
\citet{minerva-robot} & Medium & N/A & Museum & N/A & 0.70 & N/A & LIDAR & Egocentric \\
\citet{trautmanijrr} & High & P, C & Med. Room & N/A & 0.30 & N/A & Stereo & Overhead \\
\citet{truong_dynamic} & Low & P & Med. Room & N/A & N/A & N/A & LIDAR, RGB-D & Egocentric \\
\citet{tsai2020navigan} & Medium & P, C & Road & 990x670x390 & 1.00 & DD & LIDAR & Egocentric \\
\citet{wang2023navistar} & Medium & CC & L. Room & N/A & N/A & N/A &  RGB-D & Egocentric \\
\citet{wang2022feedbackefficient} & Low & CC & L. Room & N/A & N/A & N/A & RGB-D & Egocentric \\
\citet{Xie_2023} & Medium & N/A & Field & 354x354x420 & 0.50 & DD &  LIDAR & Egocentric \\
\citet{yang_online} & Low & P, R & Hallway & 700x400x500 & 3.30 & Quad & Motion Capture & Overhead \\
\citet{yao_multiple} & Low & P, C & Hallway & 354x354x420 & 0.65 & DD & LIDAR, RGB-D & Egocentric \\
\bottomrule
\end{tabular}
}
\end{scalebox}
\label{tab:literature}
\end{table*}

\subsection{Contextual Factors} \label{cf_explanation}

By contextual, we refer to factors are \textit{robot-agnostic}, i.e. they are related to the environment the robot is deployed in.

\textbf{Density}. Density is often used as a proxy for the Complexity of a scenario~\citep{trautmanijrr, Mavrogiannis2021CoreCO, sun2021trajectories}. We report it in the standard form of $agents \slash m^{2}$, although we note this does not account for the variable size of agents. Wherever a Density was not explicitly given or computable, we approximated based on supplemental footage to complete Table~\ref{tab:literature}.

\textbf{Directionality}. The directions in which humans encounter the robot contribute to the difficulty of the collision-avoidance task~\citep{wang2021corl}. We identified four cases-- Passing: Agents move parallel, paths do not intersect; Crossing: Agents move perpendicular, paths intersect; Random: Agent starts and goals are randomly sampled; Circle Crossing: Agent start and goals are sampled on opposite sides of the circumference of a circle.

\textbf{Environment Geometry}. We find that most often, evaluations take place in either hallways or medium to large office rooms, which place no constraints on the agents' movements.

\textbf{Policy Mixture}. The behavior with which each agent co-navigates affects the evolution of a scenario. Cooperative agents assume partial responsibility for collision avoidance, which simplifies the robot's task. In contrast, when rushing, being distracted, or changing intentions, humans may pose greater challenges to a robot~\citep{poddar2023crowd}. We found that most real-world studies instruct participants to navigate cooperatively, and most simulation-based studies make use of cooperative crowd simulators like Social Force~\citep{Helbing95} and ORCA~\citep{vandenberg}.

\subsection{Robot Factors} \label{rf_explanation} 

These factors refer to details of the robot hardware platform, and include the robot footprint, mass, maximum speed, and sensing capabilities, among others.

\textbf{Robot Size}. Most recent evaluations use robots with a smaller form factor than those in previous works, which allows other agents to maneuver around them more easily.

\textbf{Maximum Speed}. Consistently, the maximum speed of the robot is much lower than a natural human walking speed~\citep{poddar2023crowd}, due to hardware or safety constraints. When reviewing supplementary video material, however, we observed that this often resulted in the humans moving to avoid the robot, while the robot itself performed little to no socially aware maneuvers.

\textbf{Kinematics} Most often, we see differential drive robots used in real world experiments. This is seldom matched in simulation training and testing, in which it is frequently assumed the robot is holonomic \cite{chen-icra17, chen2020relational, Mavrogiannis2021CoreCO}.

\textbf{Sensors}. Most works use limited sensing capabilities, leading to a more challenging evaluation. We note that some of the potential negative effects of limited sensing (false detections, occlusion) are more problematic in scenarios with higher numbers of pedestrians, and might not be stress-tested in many of the evaluations we covered.

\textbf{Viewpoint}. We differentiate between \textit{overhead}, in which sensors have a bird's eye view, mitigating issues like occlusions, and \textit{egocentric}, in which sensors are on the robot.

\section{Investigating the Complexity of Social Robot Navigation Scenarios} \label{application}

We propose a definition of a \emph{Social robot navigation scenario} based on parameters that concisely capture the  contextual factors in Sec.~\ref{sec:factors-literature}. We then describe an experiment design that investigates the performance of a variety of navigation algorithms under different scenarios.

\subsection{Social Navigation Scenario}

Consider a robot navigating next to $n \geq 1$ human agents in a workspace $\mathcal{W}\subseteq \mathbb{R}^2$ with a set of static obstacles $\mathcal{W}_{obs}\subseteq \mathcal{W}$. The robot starts from an initial configuration $s_R$ and moves towards a goal $g_R$ by following a policy $\pi_R$ whereas humans are navigating from their initial configurations $s_i$ towards their goals, $g_i$ by following a policy $\pi_i$, $i\in\mathcal{N}$; agents' goals are unknown to one another. The robot occupies an area $A_R\in\mathcal{W}$, and each human occupies an area $A_{i}\in\mathcal{W}$. The objective of the robot is to reach its destination while avoiding collisions and abiding by social norms. We define a social navigation scenario as a tuple:
 \begin{equation}
    \mathcal{S}=\left(n, A_R, A_{1:n}, s_R, g_R, s_{1:n}, g_{1:n}, \pi_{1:n}, \mathcal{W}_{obs}\right)\mbox{.}
 \end{equation}
We denote by $\pi_{i}$ the true policy for agent $i$, capturing the way they make decisions based on their objectives as well as behavioral and contextual aspects of their navigation profile. 

\subsection{Experiment Design} \label{experiment_details}
We design scenarios of varying Complexity by manipulating each of the factors in Sec.~\ref{sec:factors-literature} in isolation. 

\begin{figure}[tp]
        \subfloat{%
            \fbox{\includegraphics[width=.24\columnwidth]{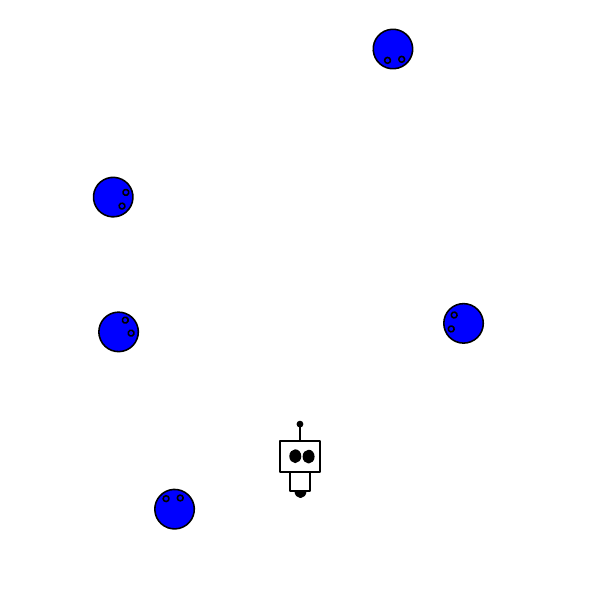}}%
            \label{subfig:a}%
        }\hfill
        \subfloat[Density]{%
            \fbox{\includegraphics[width=.24\columnwidth]{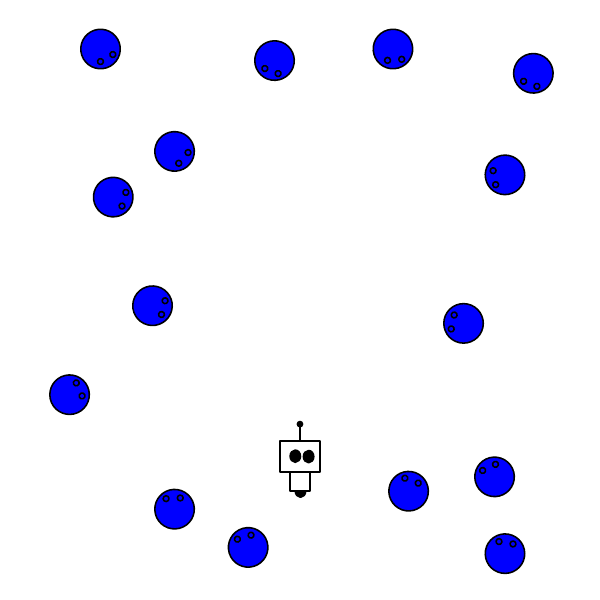}}%
            \label{subfig:a}%
        }\hfill
        \subfloat{%
            \fbox{\includegraphics[width=.24\columnwidth]{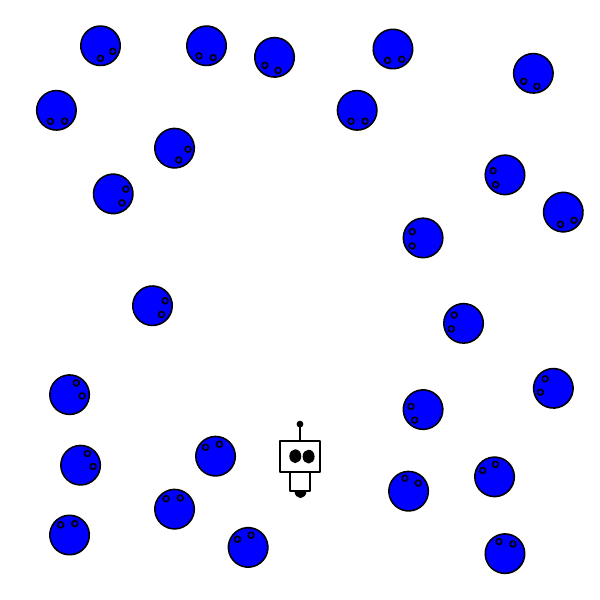}}%
            \label{subfig:b}%
        }\\
        \subfloat{%
            \fbox{\includegraphics[width=.24\columnwidth]{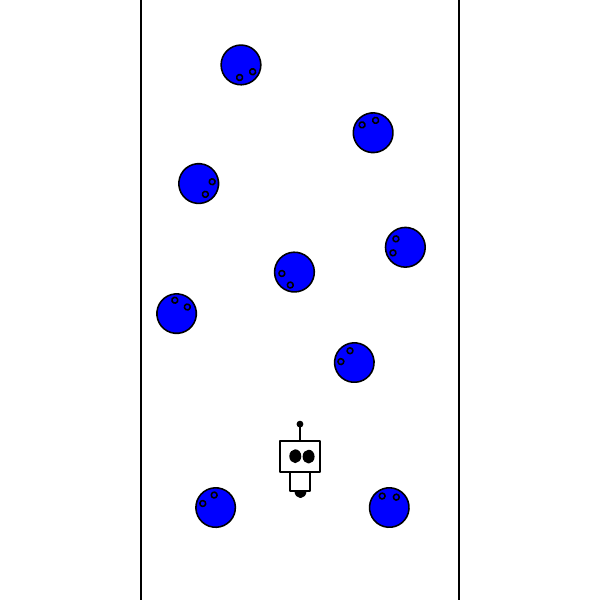}}%
            \label{subfig:c}%
        }\hfill
        \subfloat[Width]{%
            \fbox{\includegraphics[width=.24\columnwidth]{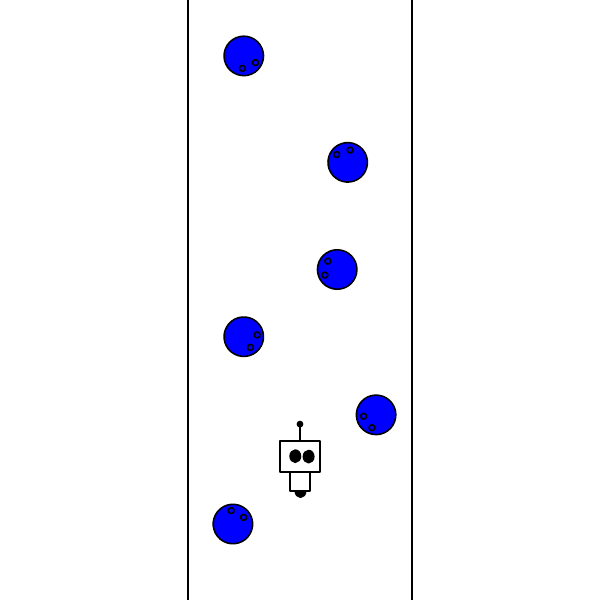}}%
            \label{subfig:a}%
        }\hfill
        \subfloat{%
            \fbox{\includegraphics[width=.24\columnwidth]{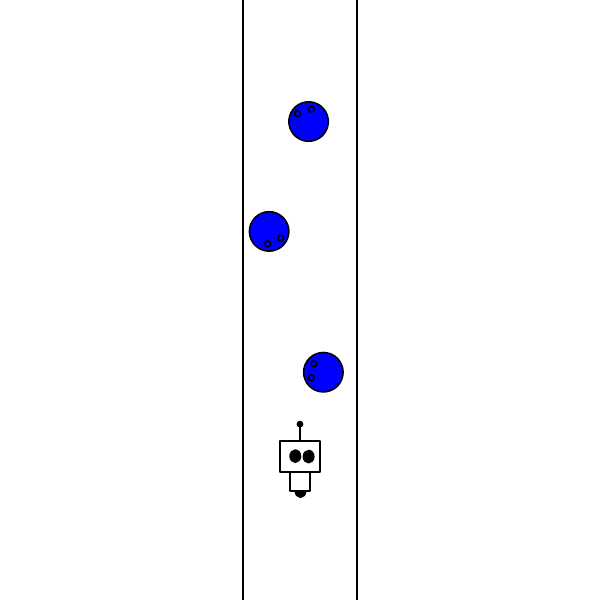}}%
            \label{subfig:d}%
        }\\
        \subfloat{%
            \fbox{\includegraphics[width=.24\columnwidth]{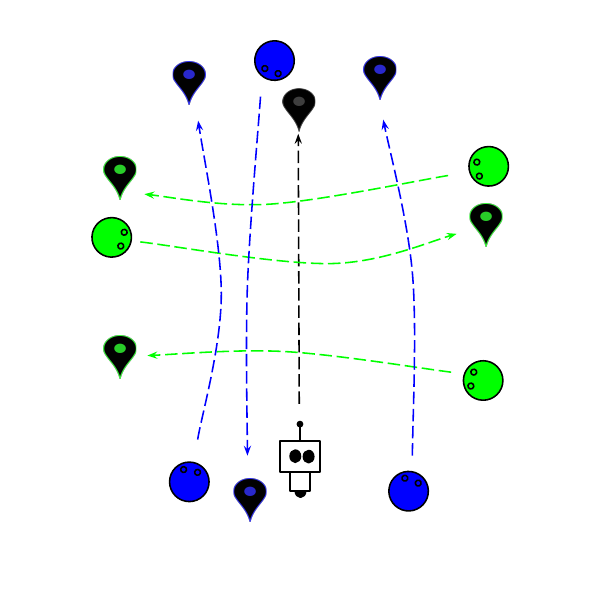}}%
            \label{subfig:c}%
        }\hfill
        \subfloat[Directionality]{%
            \fbox{\includegraphics[width=.24\columnwidth]{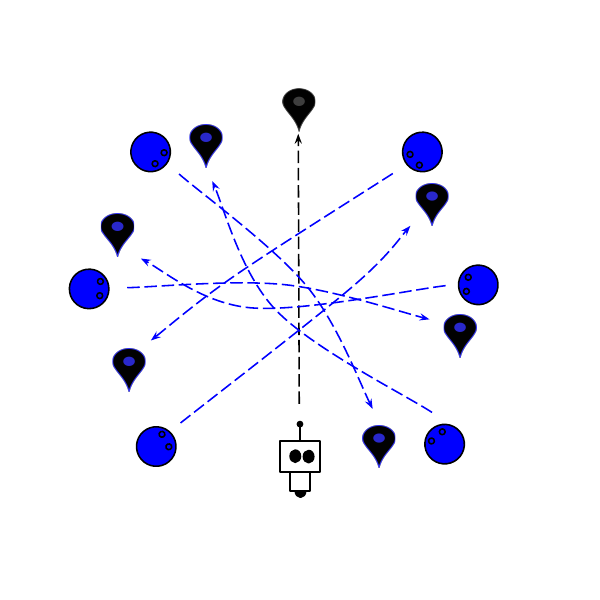}}%
            \label{subfig:a}%
        }\hfill
        \subfloat{%
            \fbox{\includegraphics[width=.24\columnwidth]{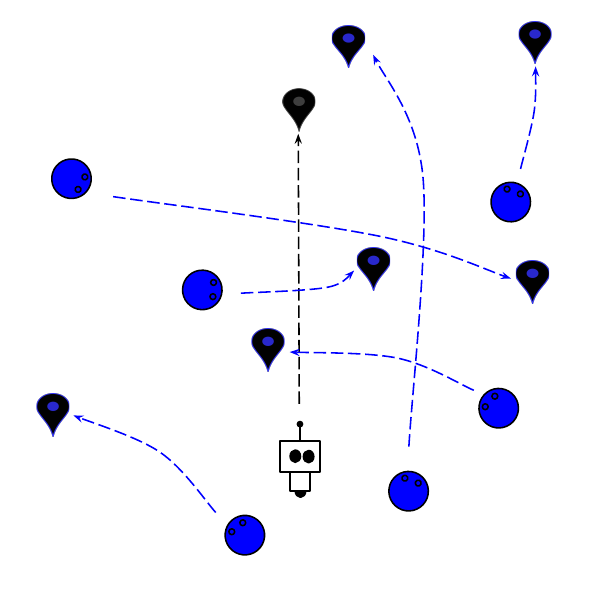}}%
            \label{subfig:d}%
        }\\
        \subfloat{%
            \fbox{\includegraphics[width=.24\columnwidth]{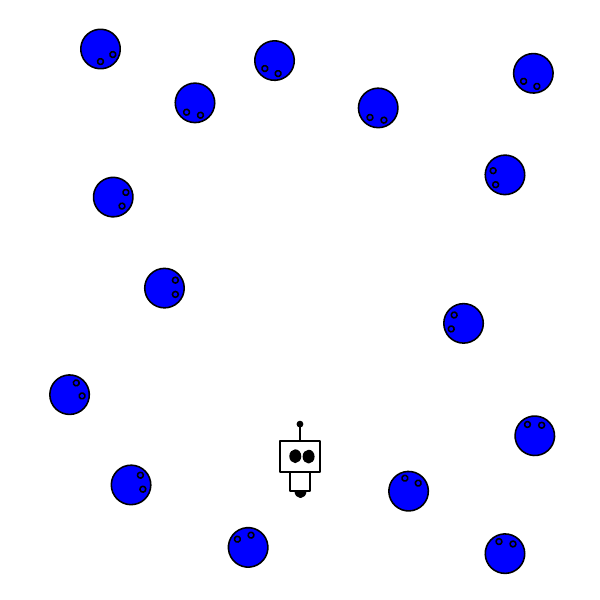}}%
            \label{subfig:c}%
        }\hfill
        \subfloat[Policy Mixture]{%
            \fbox{\includegraphics[width=.24\columnwidth]{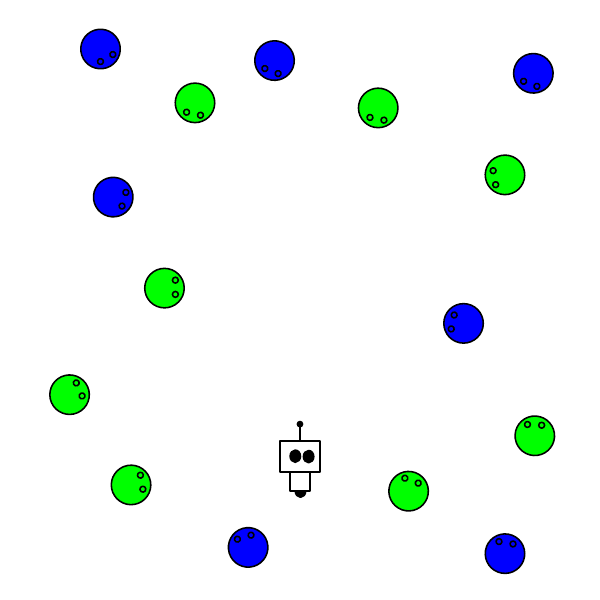}}%
            \label{subfig:a}%
        }\hfill
        \subfloat{%
            \fbox{\includegraphics[width=.24\columnwidth]{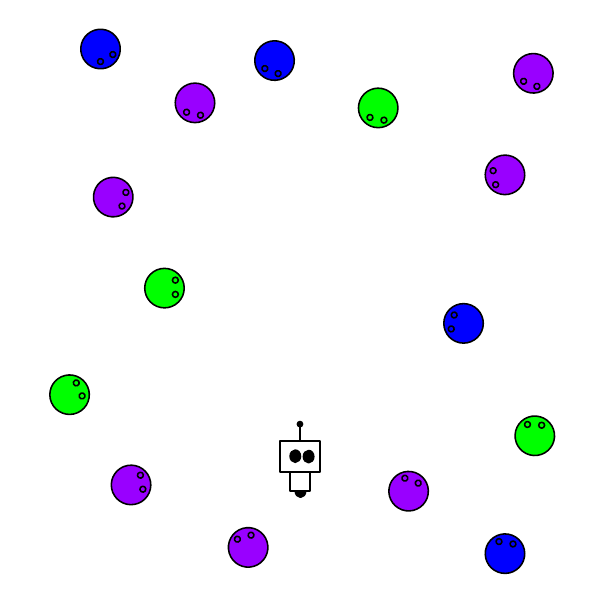}}%
            \label{subfig:d}%
        }
        \caption{Left to right: Scenarios of increasing Complexity for each of the Complexity factors considered (b-k). In the leftmost Directionality figure green agents are crossing and blue agents are passing, while different colors represent different policies in Policy Mixture.
    \vspace{-5mm}
    \label{fig:exp_configurations}}
        \label{fig:fig}
    \end{figure}

\textbf{Scenario configurations}. We first define a base scenario $\mathcal{S}_b$ with $n = 15$, $A_{R}, A_{i:n} = \pi(0.3)^2$, $s_{R} = (5, 1)$, $g_{R} = (5, 9)$, $\pi_{i:n}=$ \emph{ORCA, SFM}, $v_{pref}=1.0$ m/s, $\mathcal{W}_{obs} = \{[0, w] \times [0, l]\}^\mathsf{c} = \{[0, 10] \times [0, 10]\}^\mathsf{c}$; $s_{1:n}, g_{1:n}$ are sampled from passing and crossing. We then modify a single variable in each experiment, from low to high \textit{intensity}:

\begin{itemize}
    \item \textit{Density}: \{0.05, 0.10, 0.15, 0.20, 0.25, 0.30, 0.35\}. 
    We choose this range following prior work \cite{trautmanijrr, Mavrogiannis2021CoreCO}.
    \item \textit{Directionality}: \{Passing only, crossing only, passing and crossing, circle crossing, random start/goal\}.  We found that prior evaluations frequently involve passing scenarios with one or two humans, although many papers set up simultaneous passing and crossing scenarios. We also find that while circle crossings are frequently used in simulation to force Complex interactions between several agents \cite{chen2019crowd, chen2020relational, Mavrogiannis2021CoreCO, sun2024mixedstrategynashequilibrium}, they do not appear as often in real robot evaluations.
    \item \textit{Policy Mixture}: \{SFM only, ORCA only, Mix 1, Mix 2, Mix 3\}, where Mix 1 contains 8 ORCA and 7 SFM agents, Mix 2 contains 5 ORCA, 5 SFM, 2 CV, 3 Static agents, and Mix 3 contains 4 ORCA, 4 SFM, 4 CV, 3 Static agents, respectively. We add increasing numbers of inattentive agents to model Complex real-world scenarios in which the robot must navigate among both cooperative and uncooperative agents simultaneously \cite{matsuzaki_dwa}.
    \item \textit{Environment Width}: \{4.5, 4.0, 3.5, 3.0, 2.5, 2.0, 1.5\}. which reflects the reported Width of most hallways (e.g.,~\citep{kirby_thesis, yang_online, pacchierotti_eval}) in Table.~\ref{tab:literature}.
\end{itemize}
Fig.~\ref{fig:exp_configurations} illustrates the configurations considered in our study.

\textbf{Algorithms}. We propose to evaluate the change in Complexity of a given scenario using the performance of several navigation algorithms. Specifically, we employ:
\begin{itemize}
    \item \textit{Relational Graph Learning (RGL) \cite{chen2020relational}:} a reinforcement learning based approach that models pedestrian interactions using a graph neural network.
    \item \textit{Social GAN Model Predictive Control (MPC-SGAN):} An approach integrating a recent crowd motion prediction model \cite{gupta2018social} into an MPC framework, using the approach and implementation of~\citet{poddar2023crowd}.
    \item \textit{Constant Velocity Model Predictive Control (MPC-CV):} Identical to \textit{MPC-SGAN}, but instead using CV motion predictions for each agent.
    \item \textit{Social GAN Model Predictive Path Integral (MPPI-SGAN):} SGAN integrated into an MPPI controller.
    \item \textit{Reactive Planner (RP)}: A myopic planner that attempts to avoid collisions while navigating towards the goal.
    \item \textit{Optimal Reciprocal Collision Avoidance (ORCA)~\citep{vandenberg}}: A multiagent collision avoidance method that guarantees collision free movement among ORCA agents.
    \item \textit{Social Forces Model (SFM)~\citep{Helbing95}}: A physics-inspired model of crowd motion.
    \item \textit{Constant Velocity (CV)}: An unreactive agent that moves with constant velocity toward the goal.
\end{itemize}

\textbf{Metrics}. Based on literature~~\citep{gao-eval, Mavrogiannis2021CoreCO,Francis2023PrinciplesAG}, we use the following metrics for performance, collected from Successful trials:
\begin{itemize}
    \item \textit{Success:} The average number of trials in which the ego-agent reaches its goal without collision.
    \item \textit{Time to goal:} The average time to goal across trials.
    \item \textit{Distance to agent:} The Minimum Distance to the nearest other agent during a trial, averaged across trials.
    \item \textit{Path Irregularity \cite{guzzi_pi}:} The amount of unnecessary turning per unit path length, measured in $\frac{rad}{m}$, calculated as $\sum_{Path} \frac{Rotation - Min.\:rotation\:needed}{Path\:length}$.
\end{itemize}

\textbf{Hypotheses}. We expect that scenarios of higher Complexity will pose greater navigation challenges, and this will be reflected in significant performance drops across all algorithms. Additionally, as Density is often used as a proxy for Complexity as a whole, we anticipate that it will have the strongest correlation with observed performance drops. Based on these expectations, We formulate the following hypotheses:
\begin{itemize}
    \item \textit{H1}. Increasing the intensity of each of the four Complexity factors (Density, Directionality, Environment Geometry, Policy Mixture) independently decreases performance with respect to (w.r.t.) collected metrics.
    \item \textit{H2}. The negative correlation between Complexity and performance in \textit{H1} will be strongest for Density.
\end{itemize}

\subsection{Implementation Details}

We generate 500 scenarios for each condition within each experiment and report metrics as averages across all 500 trials. We fix the random seed to ensure each method experiences the same scenarios. To simulate continuous crowd motion, agents each have a precomputed sequence of goals sampled according to the Directionality of the scenario, and begin moving to their next goal upon reaching their current one. For MPC-SGAN and MPPI-SGAN, we used a checkpoint pretrained on the \textit{Zara} portion of the UCY dataset \cite{lerner_ucy}. RGL was retrained in 5, 10, and 15 agent scenarios with ORCA, SFM, and ORCA/SFM agent policies, however the training only resulted in a collision-avoiding policy in the 5 ORCA agent setting. For the ORCA and SFM methods, we used the CrowdNav default parameter settings. The reactive planner uses the same action space of~\citet{chen2020relational}, and selects the collision free action that minimizes its distance to the goal in the following timestep. All methods handle static obstacle collisions by setting the component of the action in the direction of the colliding obstacle to $0$. For more specific details of the training procedure for RGL, and parameter settings and tuning for the MPC and MPPI methods, as well as our enhanced version of CrowdNav, see our GitHub site at \url{https://github.com/fluentrobotics/ComplexityNav}. Videos of our experiments can be found at \url{https://youtu.be/-ir12VoSCkY}.

\subsection{Results}

\begin{figure*}
  \caption{Performance of methods across our experiments. Rows indicate experiments and columns correspond to different evaluation metrics. Each point represents the mean over 500 experiments; shaded regions indicate standard deviation. Mix 1 has 7 SFM and 8 ORCA agents. Mix 2 has 5 SFM, 5 ORCA, 2 CV, and 3 static agents. Mix 3 has 4 SFM, 4 ORCA, 4 CV, 3 static agents.}
  \hrule height .2pt depth .2pt width \textwidth
  \begin{minipage}[t]{\linewidth}
    \includegraphics[width=\linewidth]{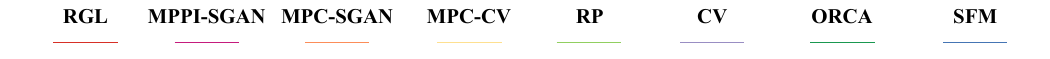}%
  \end{minipage}
   \hrule height .2pt depth .2pt width \textwidth
  \begin{minipage}[t]{\linewidth}
    \includegraphics[width=\linewidth]{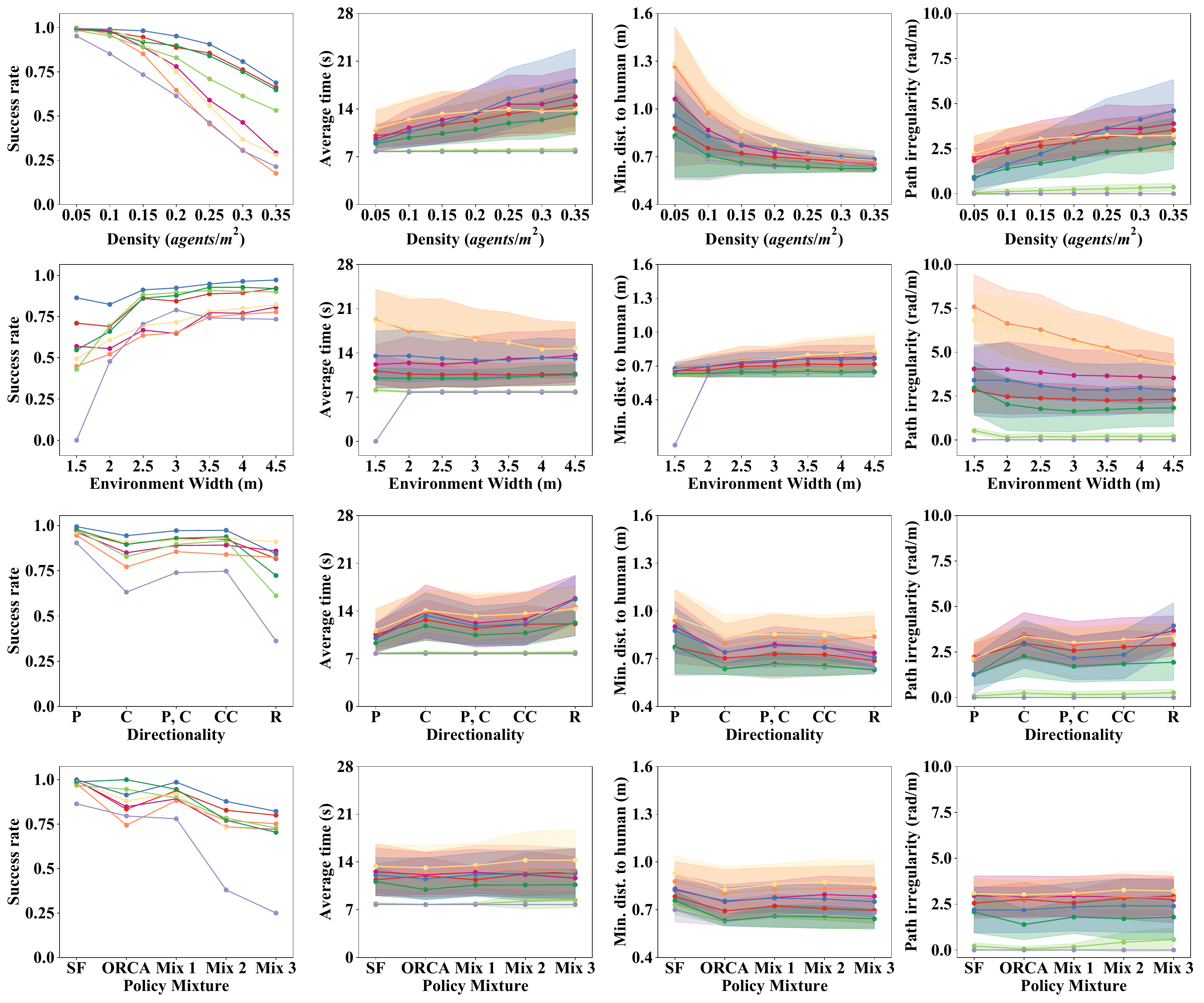}%
  \end{minipage}
  \label{complexity_results}
  \vspace{-.5cm}
  \hrule height .2pt depth .2pt width \textwidth
  \vspace{-5mm}
\end{figure*}

Fig. \ref{complexity_results} summarizes our experimental results, organized by experiment types (rows) and metrics (columns). Based on this data, we investigate the validity of our hypotheses:

\textbf{H1}. We find Density has a strong correlation with Success Rate and Minimum Distance to agent ($\rho = -0.878, -0.760$, $p < 0.001$ using Spearman's r test), and a moderate correlation with Average Time ($\rho = 0.488$, $p < 0.001$). We also find a strong correlation between \textit{increasing} Width and Success Rate ($\rho = 0.641$, $p < 0.001$), and a moderate correlation between Width and Minimum Distance ($\rho = 0.420$, $p < 0.01$). We do not find any statistically significant correlation with average Time or Path Irregularity. Regarding Policy Mixture and Directionality, we find strong and moderate correlations respectively between intensity and Success Rate ($\rho = -0.778, -0.524$, $p < 0.002$). We see no statistically significant correlation for Minimum Distance and Time in either case, and no statistically significant correlations for Path Irregularity. Thus, we find that while each factor affects at least one metric, Density and Environment Geometry appear to have the strongest correlation, giving partial support for \textit{H1}.

\textbf{H2}. We observe that the correlations between Density and other collected metrics are all stronger than those of the other factors. Thus we find support for \textit{H2}.

\subsection{Analysis}

While Density correlates strongest with Complexity, we see that the Environment Geometry, Directionality, and agent policies all have at least some correlation with performance, and thus should be considered when assessing the Complexity of a scenario. While the support for \textbf{H2} vindicates prior use of Density as a proxy for Complexity, the partial support for \textbf{H1} suggests that a more accurate picture can be obtained by analyzing the rest of the contextual factors.

Despite being trained on only 5 ORCA agent circle crossings, RGL achieves higher Success than the MPC and MPPI methods in nearly all experiments, particularly in high Density scenarios. Its Success Rate nearly mirrors ORCA (its training Policy Mixture), however with consistently higher Minimum Distances to humans. This indicates it learned similar collision avoidance behavior to ORCA, while also learning to better respect agents' personal space.

MPC-CV, MPC-SGAN, and MPPI-SGAN performance scales poorly with moderate to high Density, although we do see MPPI-SGAN scales better than MPC-SGAN, showing that a stronger controller implementation does indeed lead to better performance (albeit with slightly lower Minimum Distances). The steep decrease for all three methods, however, indicates that having an inaccurate prediction model becomes increasingly problematic as the number of agent trajectories predicted increases, even with a more robust controller.

In nearly all experiments, RP and ORCA policies had low average Time, while still maintaining comparable or better Success rates to other methods. Even the CV agent was moderately Successful in lower Complexity scenarios, which matches prior experimental outcomes~\citep{Mavrogiannis2021CoreCO}. SFM's high Success Rates and Minimum Distances indicate that with proper tuning it could be viable as a local controller for a socially navigating robot, although more must be done to evaluate whether its effectiveness transfers out of simulation.

We do however see that the MPC methods maintain the highest Minimum Distance, while ORCA, RP, and CV have the lowest Minimum Distances. Additionally, the MPC methods, with the exception of the Environment Width experiments, maintain comparable Path Irregularity to ORCA, SFM, and RGL, showing they can maintain distance with comparably smooth paths. Thus we see that while the reactive and CV methods (with the exception of SFM) achieve efficient, collision-free navigation among cooperative agents, they are worse at respecting the personal space of others compared to predictive methods.

We obtained evidence that increased heterogeneity of scenarios is not by itself an indicator of increased Complexity. Qualitatively, we observed that passing scenarios are generally easier to navigate than crossing, and SFM agents are much more subservient than ORCA. An implicit hypothesis in our experimental ordering was that a mixture of two Directionalities or Policy Mixtures would be more Complex to navigate than either individually. We instead see improvement in the ORCA/SFM and Passing/Crossing scenarios compared to ORCA and Crossing only, which suggests the combinations are actually less Complex.
\section{Discussion}

\textbf{Conducting high-Complexity evaluations}. Evaluations in the literature are often conducted under low-Density settings (Table~\ref{tab:literature}). Our experiments demonstrate that most algorithms handle those well but experience significant performance reductions as Density increases. Similarly, while many works contain passing-only experiments, in our study, Passing has the highest Success rates and Minimum Distances to agents, indicating it is the easiest to navigate. Additionally, while many works consider Medium-to-Large rooms, these appear to be the simplest to navigate (Fig.~\ref{complexity_results}). Finally, nearly all evaluations in recent literature focus on scenarios with exclusively cooperative agents, which our experiments show are easiest. These practices hinder the extraction of helpful insights; for example, had we not manipulated the Complexity factors towards the upper extremes, the severe performance drops at high Density experienced by MPC-CV, MPC-SGAN, and MPPI-SGAN would not have been identified. These observations suggest that the social navigation community should shift towards studying scenarios of higher Complexity to move beyond the frontier. Specifically, more evaluations in more geometrically-constrained, highly-mixed and high-traffic environments, similar to many context-rich public domains should be conducted.

\textbf{Handling test-time distribution}. The high Success rates of CV and RP suggest that in lower-Complexity regimes, it might be sufficient to use non-reactive classical planning techniques. As the Complexity increases, all algorithms experience a steep performance decline, the magnitude of which serves to indicate how far out of distribution an algorithm operates. Our evaluation (Fig.~\ref{complexity_results}) captures the sensitivity of data-driven approaches to their training distribution. For instance, we see that RGL, and MPC/MPPI-SGAN experience a substantial performance decline as the test-time Density moves away from the training/tuning Density ($0.05, 0.10$ $agents/m^2$). In practice, data-driven approaches will often face distribution shift at deployment, leading to unexpected or unsafe behavior, as demonstrated by the performance drop experienced by RGL and the MPC approaches in scenarios of different Complexity than their training data. Thus, to safely and effectively deploy social navigation algorithms, it will be important to rigorously characterize the scenarios in which they can be expected to perform effectively, and which scenarios are out of distribution and may cause performance degradation.

\textbf{Reproducing Complexity}. In our literature review, we found that the amount of details provided varied substantially among papers, leaving us to approximate values based on figures or supplementary material, when available. We believe that sharing more precise details related to the Complexity factors we identified will be an important step towards better reproducibility in social navigation.

\textbf{Simulating complex settings}. While realistically simulating pedestrian-robot interactions is challenging~\citep{fraichard2020, Mavrogiannis2021CoreCO}, we view it as an essential tool for testing high-Complexity settings that are difficult to safely replicate in the real world. This requires revisiting conventional assumptions, such as that humans are non-reactive to the robot~\citep{mun2023occlusionaware, liu2023intention, wang2022feedbackefficient, matsuzaki_ddrl, chen-icra17, chen2020relational}, which is unrealistic since the robots used (see Table~\ref{tab:literature}) generally have a large enough footprint to be observed. Thus, simulation should focus on \emph{visible} robot settings, leveraging metrics and considerations of users' perceptions~\citep{social-momentum-thri, singamaneni_bench,walker2021corl} to compare algorithms' social performance.

\textbf{Limitations}. Our analysis is a step towards understanding the Complexity of social navigation scenarios; an important next step would be to validate our observations with real-world experiments. While our evaluation involves standard metrics~\citep{Mavrogiannis2021CoreCO,Francis2023PrinciplesAG}, it lacks a user-centered perspective. Future work will incorporate metrics mapping robot behavior to users' impressions~\citep{dadvar2022joint,social-momentum-thri,walker2021corl}. We tested algorithms for which we found open-source implementations but additional algorithms could be included. We strived for a fair comparison by tuning all algorithms w.r.t. the same criteria. While tuning can always be improved, the key takeaway remains: there is a steep drop in performance as Complexity increases. Additionally, we acknowledge that our simulator makes the unrealistic assumptions of perfect sensing and constant-sized circular agents, and these could be improved in future implementations. Furthermore, our results are affected by our choice of ORCA and SFM to simulate cooperative agents, where other algorithms may have led to different results. While these assumptions reduce the realism of our experiments, they allow us to minimize the effect of extraneous factors on the relationship between the Complexity factors and performance. Future work will investigate the robot-related factors of kinematics, sensors, and viewpoints, and can explore the effects of additional real-world considerations.

\footnotesize
\bibliography{references}
\bibliographystyle{IEEEtranN}


\end{document}